\documentclass[runningheads]{llncs}

\usepackage[T1]{fontenc}
\usepackage{cite}
\usepackage{amsmath,amssymb,amsfonts}
\usepackage{algorithmic}
\usepackage{graphicx}
\usepackage{subfigure}
\usepackage{textcomp}
\usepackage{xcolor}
\usepackage{comment}
\usepackage{hyperref}
\usepackage{booktabs}
\usepackage{caption}
\usepackage{threeparttable}
\usepackage{marvosym}
\hypersetup{hidelinks}

\def\BibTeX{{\rm B\kern-.05em{\sc i\kern-.025em b}\kern-.08em
		T\kern-.1667em\lower.7ex\hbox{E}\kern-.125emX}}

\usepackage{multirow}
\usepackage[linesnumbered,ruled,vlined]{algorithm2e}

\newcommand{\M}{Word-Graph2vec}
\newcommand{\SG}{Skip-Gram}
\newcommand{\FT}{FastText}

\begin{document}

\title{\M: An efficient word embedding approach on word co-occurrence graph using random walk technique}
\titlerunning{\M}

\author{Wenting Li\inst{1}\and
	Jiahong Xue\inst{2}\and
	Xi Zhang\inst{2}\and Huacan Chen\inst{2}\and Zeyu Chen\inst{2}\and Feijuan Huang \textsuperscript{\Letter} \inst{3} \and Yuanzhe Cai \textsuperscript{\Letter} \inst{2}}

\authorrunning{Wenting Li et al.}

\institute{Shenzhen Technology University, Shenzhen, Guangdong, China\\ \email{liwentingnwz@gmail.com} \and Shenzhen Technology University, Shenzhen, Guangdong, China \and Shenzhen Institute of Translational Medicine, Shenzhen Second People’s Hospital, The First Affiliated Hospital of Shenzhen University, Shenzhen, Guangdong, China}

\maketitle             

\begin{abstract}
Word embedding has become ubiquitous and is widely used in various natural language processing (NLP) tasks, such as web retrieval, web semantic analysis, and machine translation, and so on. Unfortunately, training the word embedding in a relatively large corpus is prohibitively expensive. We propose a graph-based word embedding algorithm, called~\M, which converts the large corpus into a word co-occurrence graph, then takes the word sequence samples from this graph by randomly traveling and trains the word embedding on this sampling corpus in the end. 
We posit that because of the limited vocabulary, huge idioms, and fixed expressions in English, the size and density of the word co-occurrence graph change slightly with the increase in the training corpus. 
So that~\M~has stable runtime on the large-scale data set, and its performance advantage becomes more and more obvious with the growth of the training corpus. 
Extensive experiments conducted on real-world datasets show that the proposed algorithm outperforms traditional Word2vec four to five times in terms of efficiency and two to three times than~\FT, while the error generated by the random walk technique is small.

\end{abstract}

\section{Introduction}

% word embedding is important. 
%Word embedding means embedding words into continuous real space as vectors, allowing words with similar meanings to have a similar representation. It is widely used in modern natural language processing (NLP) tasks, including semantic analysis \cite{yu2017refining}, sentiment analysis \cite{faruqui2014retrofitting}, information retrieval\cite{manning2010introduction}, machine translation\cite{chen2017neural}, question answering system\cite{hao2017end} and so on. 

Word embedding is widely used in modern natural language processing (NLP) tasks, including sentiment analysis \cite{faruqui2014retrofitting}, web retrieval\cite{manning2010introduction}and so on. 
Current word embedding methods, such as Word2vec\cite{mikolov2013efficient} and Glove\cite{pennington2014glove}, rely on large corpora to learn the association between words and obtain the statistical correlation between different words so as to simulate the human cognitive process for a word. 
% The statistical relationship between words in natural language processing tasks is called context, and its goal is to predict the words in the local context to the greatest extent. 
% it is a time-consuming process. ,  
The time complexity of these two approaches is $O(|N| log(|V|))$, where $|N|$ is the total corpus size, and $|V|$ is the size of vocabulary, which clearly indicates that the runtime of these two training approaches increases linearly as the size of corpus increases.
% big data needs the new approaches.
In the era of big data, how to speed up these existing word embedding approaches becomes increasingly essential.

In this paper, we address the problem of efficiently computing the word embedding on a large-scale corpus. 
Our solution is also established on a word co-occurrence graph, Then text sampling by random walk traveling on the graph. 
In detail, we intend to go one step further and propose a graph-based word embedding method called~\M.
This approach contains three steps. 
First,~\M~uses word co-occurrence information to construct a word graph whose each word as a node, whose edges represent co-occurrences between the word and its adjacency node, and whose edge direction represents word order. 
Second,~\M~performs random walk traveling on this word graph and samples the word sequences.
Third, skip-gram~\cite{mikolov2013efficient} has been applied to these sampling word sequences to gain the final word embedding. 

% Explain the advantage of our approach. 
The main advantage of~\M~is the performance on the large-scale corpus. 
Because of the limited vocabulary~\footnote{The vocabulary of the New Oxford Dictionary is around 170,000, but some of the words are old English words, so that in actual training, the word co-occurrence graph contains about 100,000 to 130,000 nodes.}, the number of nodes and density of the word co-occurrence graph change slightly with the increase of training corpus. 
So that~\M~has stable runtime on the large-scale data set, and its performance advantage becomes more and more obvious with the growth of the training corpus. 
Parenthetically, noted that adding more training corpus only results in adjusting the edge weights.
Therefore, the time consumed by training the model increases slowly as the corpus increases.

\section{Related Works}
\label{sec:related_work}

We categorize existing work related to our study into two classes: word embedding approaches and graph embedding approaches.

\noindent \textbf{Word Embedding Approaches:} 
One of the most prominent methods for word-level representation is Word2vec~\cite{mikolov2013efficient}. 
So far, Word2vec has widely established its effectiveness for achieving state-of-the-art performances in various clinical NLP tasks. 
GloVe~\cite{pennington2014glove} is another unsupervised learning approach for obtaining a single word's representation. 
%It is a statistical model aggregating a global matrix factorization and a local context window. The learning relies on dimensionality reduction on the co-occurrence count matrix, which is based on how frequently a word appears in a context. 
Different from Word2vec and GloVe, FastText~\cite{si2019enhancing} considers individual words as character n-grams. Words actually have different meanings in different contexts, and the vector representation of the two model words in different contexts is the same. 
%ELMO~\cite{DBLP:journals/corr/abs-1802-05365} is optimized for this. ELMO can learn the complex characteristics of word usage and the changes of these complex usages in different contexts. 
However, the structure of these pre-training models is limited by the unidirectional language model (from left to right or from right to left), which also limits the representation ability of the model so that it can only obtain unidirectional context information. 

%BERT~\cite{devlin2018bert} uses Masked Language Model (MLM) and Next Sentence Prediction(NSP) for pre-training and uses deep bi-directional transformer~\cite{vaswani2017attention} components to build the whole model. 
%Hence, it finally generates a deep two-way language representation that can integrate left and right context information.

%All these methods achieve satisfactory results in speed and accuracy, but when encountering a large-scale corpus, the training time cost will increase significantly. 

\M~model makes the training time in a stable interval with the increasing of the size of the corpus and also ensures the accuracy for various NLP tasks.

\noindent \textbf{Graph Embedding Method:} In graphical analysis, traditional machine learning methods usually rely on manual design and are limited by flexibility and high cost. Based on the idea of representational learning and the success of Word2vec, Deepwalk\cite{perozzi2014deepwalk}, as the first graph embedding method based on representational learning, applies the~\SG~model to the generated random walk. Similarly, inspired by Deepwalk, Node2vec\cite{grover2016node2vec} improves the random walking mode in Deepwalk.  Node2vec introduces a heuristic method, second-order random walk. Considering the difference between the linear structure of the text and the complex structure of graphics, our model adopts the idea of Node2vec for node learning
% which uses two super parameters $p, q $ and weight parameter $ \alpha $ to control the walking strategy. Therefore, the generated sequence is a combination of DFS and BFS. 

%Modeling text as a word-connected association network (word graph)~\cite{rousseau2013graph} is a good exploration direction in linguistics.
%Word graph can seamlessly integrate grammar, semantics, and other information in text language into the complex structure of the graph. 
%The advantage of studying word graphs instead of studying the whole text is that we can easily express language's non-linear and non-hierarchical structure.
%These graph embedding techniques have been surveyed and used to sample the word sequences on the word graph, but these approaches do not consider the weight of nodes in the sampling process. 
%Therefore, in our~\M~approach, the weight of each node has been set, and the number of random walks of each node is decided on its weight in the corpus, which is more in line with the characteristics of the text.
\section{~\M~algorithm}
\label{sec:algorithm}

% In this section, the motivation of \M~will be illustrated, and a summary of the whole process will also be given. And the part of our improvement is explained in detail. We also discuss the setting of some parameters in the algorithm and the complexity.

\subsection{Motivation}
\label{sec:motivation}

% Yuanzhe: These sentences should add in the related work section. 

% Word2vec learns the distributed representation of words from a large number of unmarked texts, which not only contains a large amount of information, but also can be migrated to various downstream tasks. Similarly, the network has many paths, connecting each node into a line. These connections also contain the relationship between nodes, just like the relationship between words in a sentence. Therefore, the author of Node2vec tries to treat these node sequences as sentences and trains them with the Word2vec method to get a better vector representation of the node. No matter what kind of node embedding, it is necessary to obtain a sequence first, and many algorithms also work hard on how to obtain this sequence. From the perspective of intuitive feeling, a random walk on the graph is obviously a more appropriate way. In fact, it is also a kind of random sampling.

%Yuanzhe: Graph description is very popular.
Word graphs, extracted from the text,  have already been successfully used in the NLP tasks, such as information retrieval~\cite{blanco2012graph}and text classification~\cite{hassan2007random}.  
The impact of the term order has been a popular issue, and relationships between the terms, in general, are claimed to play an important role in text processing. 
For example, the sentence \textit{``Lily is more beautiful than Lucy''} is totally different from the sentence \textit{``Lucy is more beautiful than Lily''}.
This motivated us to use a word co-occurrence graph representation that would capture these word relationships. 
 
Training the word embedding on a word co-occurrence graph is an efficient approach. 
First, the number of nodes in this word co-occurrence graph is not large.  
According to lexicographer and dictionary expert, Susie Dent, ``the average active vocabulary of an adult English speaker is around 20,000 words, while his passive vocabulary is around 40,000 words."
Meanwhile, even for the large En-Wikipedia data set, the total number of word graph nodes is around 100,723, and these word graphs are highly sparse(see Table~\ref{tab:data_statistic}).  
%Therefore, these graph embedding approaches, such as DeepWalk~\cite{perozzi2014deepwalk}, Node2Vec~\cite{grover2016node2vec}, Line~\cite{tang2015line} etc., can be efficiently used for these co-occurrence graphs.    
%Second, these word graphs are highly sparse(See Tbale ).  
As the size of the training corpus increases, the weight on edges will change a lot, but the size of nodes and graph density do not have significant alterations. 
Thus, this pushes us to propose our approach,~\M, to address this word embedding issue. 

\subsection{Framework of~\M}
\label{sec:framework}

\begin{figure}[h]
	\centering
	\includegraphics[width=\linewidth]{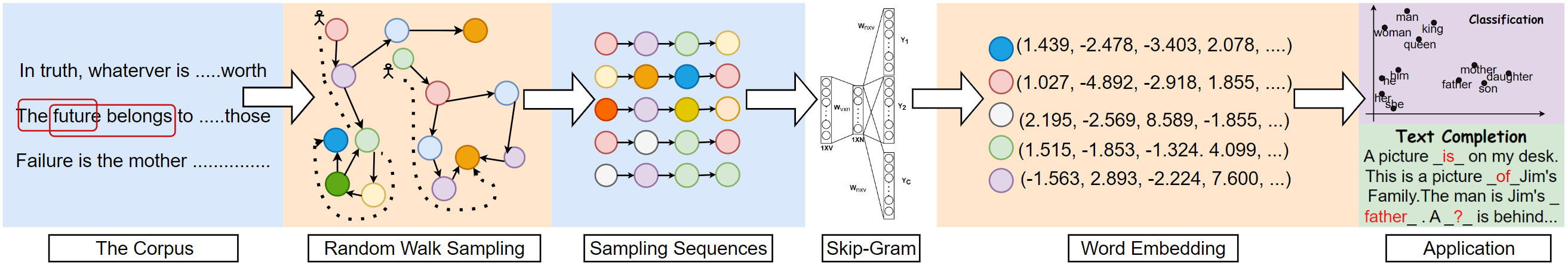}
	\caption{The overall framework for Word-Graph2vec algorithm. }
	\label{fig:framework}
\end{figure}

Figure~\ref{fig:framework} shows the framework of~\M.
%, consisting of three steps.
%First, the word co-occurrence graph is built from the training corpus. 
%Second, the random walk technique has been applied to generate the word sequences from the word graph.
%Third, the sequence set, as a new corpus, is used to train the word embedding by~\SG~algorithm.
%The following sections will explain them in detail.

\vspace{-0.3in} 

\subsubsection{\textbf{Generate word co-occurrence  graph}}
\label{sec:generate_word_relation_graph}

\begin{figure}
	\centering
	\subfigure[An example sentence. A solid arrow represents a new directed edge while a dashed arrow an already existing one in the graph. The edges are drawn from or to word ``doing".]{\label{fig:subfig:a}
		\raisebox{0.6\height}{\includegraphics[width=0.4\linewidth]{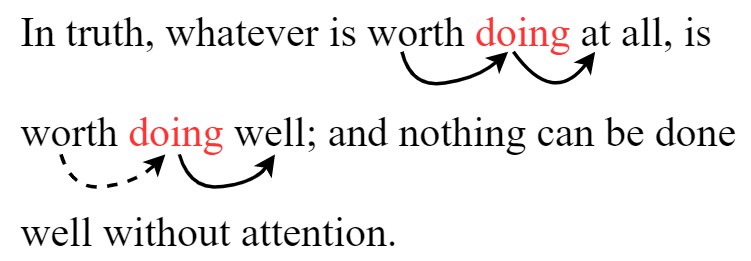}}}
	\hspace{0.01\linewidth}
	\subfigure[The word co-occurrence graph. The red dotted line describes a random walk with ``whatever" as the starting node and ``attention'' as the endding node. ]{\label{fig:subfig:b}
		\includegraphics[width=0.4\linewidth]{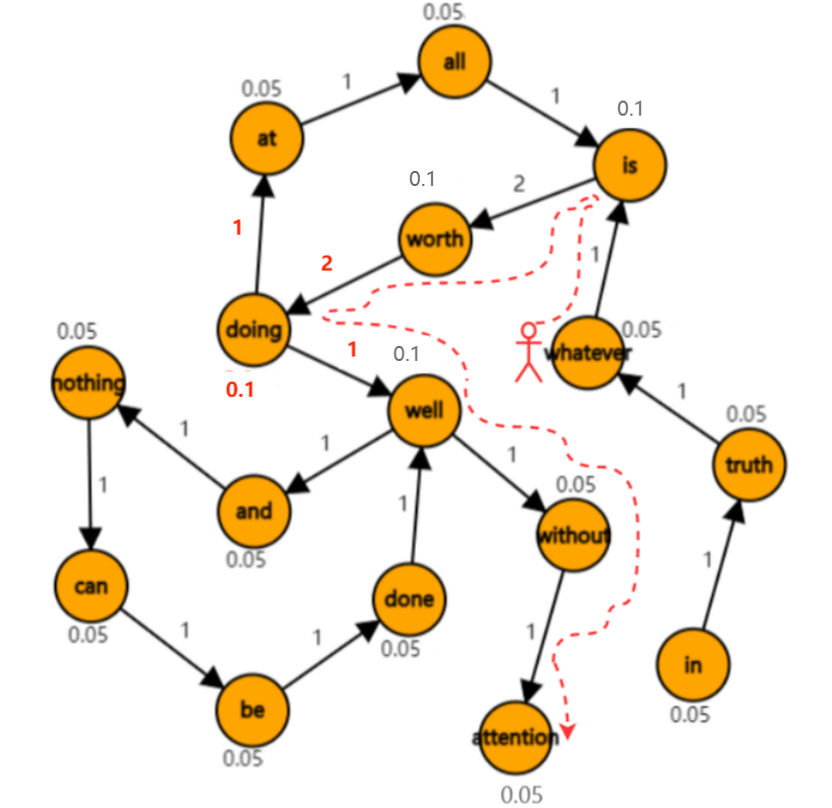}}
	\vfill
	\caption{An example of word co-occurrence graph creating from a sentence in Philip Dormer Stanhope's letter.}
	\label{fig:subfig}
\end{figure}

A textual document is presented as a word co-occurrence graph that corresponds to a weighted directed graph whose vertices represent unique words, whose edges represent co-occurrence between the words, and whose edge direction represents word order. 
An example of graph creation is given in Figure~\ref{fig:subfig:a}. 
The source text is an extract of a sentence from Philip Dormer Stanhope's letter, ``In truth, whatever is \textbf{worth doing at} all, is \textbf{worth doing well}; and nothing can be done well without attention." 
%An edge is drawn between a word and its following word, and a solid arrow represents a new directed edge, while a dashed arrow is an already existing one in the graph. 
%For clarity purposes, the edges are created from the vertex corresponding to ``doing".
Figure~\ref{fig:subfig:b} corresponds to the resulting weighted directed graph where each vertex represents a unique word, and each edge is a co-occurrence of the two words.

\noindent \textbf{Weight on edges:} The number of simultaneous occurrences of these two words in the text is used as the weight on edge. A weighted adjacency matrix $W$ is used to store the edge weights, and $W_{v,x}$ is the weight from node $v$ to node $x$. 

\begin{equation}
	W_{v,x}=\sum co\mbox{-}occurrence(v,x) 
\end{equation}  
where $ co\mbox{-}occurrence(v,x) $ is the number of times that the words $v$ and $x$ appear together from left to right in the same order. 
%For example, in Figure~\ref{fig:subfig:b}, the number of occurrences of word pairs \textit{worth} and \textit{doing} is 2, so $ W_{worth, doing}$ is equal to 2. Similarly, we get the relationship between each word in the whole sentence, and the generated word graph is shown in Figure~\ref{fig:subfig:b}.

%\begin{figure*}[h]
%	\centering
%	\includegraphics[width=\linewidth]{figures/text_arrow.png}
%	\caption{An example of graph creating from a short sentence. A solid arrow represents a new directed edge while a dashed arrow an already existing one in the graph.}
%	\label{fig:sentence_example}
%\end{figure*}
%
%\begin{figure}[h]
%	\centering
%	\includegraphics[width=\linewidth]{figures/word_graph.png}
%	\caption{An example of a graph with only one sentence}
%	\label{fig:graph_example}
%\end{figure}

\noindent \textbf{Weight on nodes:} 
% Obviously, if the nodes are strongly connected with a large weight, that is, when the nodes gather together, they often appear simultaneously in a short random walk\cite{musial2019analysis}.
The weight of the node is used to determine the sampling times in the random walk process.
We take probability $ PW_{v} $ as a representation of the importance of the word $v$ in the whole corpus, that is, the weight of nodes in the graph.
This paper uses the terms frequency(TF) and inverse document probability(TF-IDF) to set the word weights $PW$.

%First, intuitively, the term frequency (TF) can be applied to measure the importance score. 
%
%\begin{equation}
%	\label{equ:tf}
%	PW_{v}=\frac{n_{v}}{\sum_{k} n_{k}} 
%\end{equation}
%where $PW_{v}$ is the important score for word $v$,  $n_{v}$ is the number of occurrences of the word $v$, and  $\sum_{k} n_{k}$ is the number of occurrences of all the words. 
%Taking Figure~\ref{fig:subfig:b} as an example, the total number of words is equal to 20, then $ PW_{doing} = \frac{2}{20} = 0.1$.

%Second, to minimize the weight of frequent terms while making infrequent terms have a higher impact, inverse document frequency (IDF) is also used in our importance score. Thus, we have the following equation. 
%
%\begin{equation}
%	\label{equ:tfidf}
%	PW_{v} = \frac{n_{v}}{\sum_{k} n_{k}} \times  \log \frac{|C|}{|\{d: v \in d\}|}
%\end{equation}
%where  $\log \frac{|C|}{|\{d:v \in d\}|}$ is the equation to calculate the IDF score, 
%$|C|$ is the number of documents in the corpus, $d$ is the document in the corpus, and $|{d:v \in d}|$ is the number of documents that contain the word $v$.
% However, because the training data sets (Text8, 1b words benchmark, and En-Wikipedia) for word embedding is a long word sequence, the window size, which is set to 200 in our experiment, is used to identify the different documents. In other words, the words in the same window mean these words are in the same document; otherwise, these words are in different documents.      

\subsubsection{\textbf{Sampling word sequences by random walk}}
\label{sec:sampling_word_sequence}

This random walk sampling process~\cite{wang2020novel} iteratively explores the global structure network of the object to estimate the proximity between two nodes. Generate a random walk from the current node, and select the random neighbors of the current node as a candidate based on the transition probability on the word co-occurrence graph. 
%For example, in Figure~\ref{fig:subfig:b}, one random walking sampling sequence is ``whatever, is, worth, doing, well, without, attention", which is starting from ``whatever" to ``attention".  
%In addition, these random walk sampling word sequences have been collected as the training corpus for learning the word embedding. Two points are worth highlighting in this sampling process. 

First, the different nodes, as distinct from the same selection probability for graph embedding, should have different probabilities to be selected as the starting point. We use the probability sampling method based on the word weight is applied to select the rooted node.  
For example, the sampling times of node $v$ as the starting node are shown in Equ~\ref{equ:ru}:

\begin{equation}
	number\_walks(v) = \left \lfloor total\_walks \times PW_{v} \right \rfloor
	\label{equ:ru}
\end{equation} 

Where $total\_walks$ is the total number of random walk sampling, and $PW_{v}$ is the weight of word $v$.
% For example, let $total\_walk = 200$, $ number\_walks(\text {doing})$ should be $200 \times \frac{2}{20} = 20$. That means there are 20 word sequences which is starting from word ``doing". 

Second, considering the shuttle between two kinds of graph similarities (homophily and structural equivalence), Node2vec~\cite{grover2016node2vec} is selected as the graph sampling approach. 
%To set the different $p$ and $q$ values, Node2vec can select the different walking strategies (breadth-first or depth-first) to travel the graph. 
Then, according to the the $2^{nd} $ order random walk transfer probability of the Node2vec model, the random neighbours of the current node is selected as the candidate node.
The sampling sequence of a node is determined by simulating several biased random walks of fixed length $l$. 

\subsubsection{\textbf{Learning embedding by~\SG~Model}}
\label{sec:learning_embedding_with_skip_gram_model}

%Word2vec can utilize either of two model architectures to produce a distributed representation of words: continuous bag-of-words (CBOW)~\cite{mikolov2013exploiting} or continuous~\SG~\cite{mikolov2013efficient}. 
%In the continuous bag-of-words architecture, the model predicts the current word from a window of surrounding context words. 
%The order of context words does not influence prediction (bag-of-words assumption). 
%In the continuous~\SG~architecture, the model uses the current word to predict the surrounding window of context words. 
%In paper \cite{menon2020empirical}, CBOW is faster while ~\SG~ does a better job for NLP tasks.
%Thus,~\SG~algorithm is used to learn the word embedding from these sampling corpora. 

After obtaining a much smaller set of walking sequences compared to the original corpus, we used the Skip-Gram model to learn the final word embedding.

\subsection{\M~Algorithm}
\label{sec:m_algorithm}

\M~algorithm is presented in Algorithm~\ref{alg:word_graph2vec}.

%The input is the large-scale corpus $C$. Our task is to compute the word embedding $\Phi$ from these corpus $C$.   
%\M~algorithm contains seven arguments. 
%The first argument $C$ is our input corpus. 
%The second and third parameters are the number of sampling $n$ and walk length $l$. 
%These two parameters are used to control the number of sampling sequences and the length of the sampling sequence. 
%As we mentioned before, the fourth and fifth parameters $q$ and $p$ are used to adjust the random walk strategy.
%The sixth and seventh parameters, context size $k$ and dimension $d$, are used in the~\SG~algorithm. 
%The context size $k$ is used to adjust the size of the sliding window, which is used to identify whether or not these two words are in the same context. 
%Dimension $d$ is used to set the dimension of embedding. 
%This algorithm, first, constructs the word co-occurrence graph by scanning the whole corpus. 
%Then, random traveling on the word co-occurrence graph generates the bunches of word sequences. 
%Finally, word embedding has been learned from these sequences.          

\begin{algorithm}[tb]
	\caption{\textit{\M~Algorithm}}
	\label{alg:word_graph2vec} 
	
	\textbf{Input}: A processed corpus $C$, The number of sampling $n$, Walk length $l$, $p$, $q$
	Context size $k$,  Dimension $d$\\
	\textbf{Output}: Word vector representations $\Phi$
	\begin{algorithmic}[1] %[1] enables line numbers
		
		\STATE $ W \leftarrow \emptyset $   \tcp*[h]{Adjacency Matrix} \;
		\STATE $ PW \leftarrow  \emptyset$   \tcp*[h]{Weight on nodes} \;
		
		\STATE \For {$ i = 1 $ \KwTo $ Size(C) $}{
			\For {$ word $ \KwTo $C_{i}$}{
				$ W_{word,word+1} \leftarrow W_{word,word+1} + 1$ \
				$ PW_{word} \leftarrow PW_{word} + 1$ \
			}
		}
		
		\STATE $\pi$ $\leftarrow$  PreprocessModifiedWeights$(W, p, q)$   \;
		
		\STATE $ G^{\prime}=(V, E, \pi, PW) $\;
		
		\STATE $SC  \leftarrow \emptyset $  \tcp*[h]{SC is the result set of random walk sampling.}
		
		\STATE \For{all nodes $u \in V $}{
			\For{$iter = 1$ \KwTo $n\frac{PW_u}{\sum_{k}PW_k}$}{
				$ word\_sequence \leftarrow \emptyset$ RandomWalkSampleProcess $ (G^{\prime}, u, l) $\;
				Append $ word\_sequence $ to $ SC $
			}
		}
		\STATE $\Phi$ = Skip-Gram$ (k, d, SC) $\;
		\STATE \Return $\Phi$\;
		\BlankLine 
		
%		\STATE \textbf{function} RandomWalkSampleProcess(Graph $ G^{\prime}=(V, E, \pi, PW) $, Start node $u$, Length $l$)\;
%		\STATE Initialize $word\_sequence$ to $ \left[u \right] $ \;
%		\STATE \For {$walk\_iter = 1$ \KwTo $l$}{
%			$curr \leftarrow walk \left[-1\right]$\;
%			$V_{curr} \leftarrow $ GetNeighbors$ (curr,G^{\prime}) $\;
%			$ s \leftarrow $ AliasSample$ (V_{curr},\pi) $\;
%			Append $ s $ to $word\_sequence$\;
%		}
%		\STATE \Return $word\_sequence$\;
	\end{algorithmic}
\end{algorithm}

The algorithm first initializes variables (Lines 1-2). 
We scan all the corpus once and build the word co-occurrence graph (Lines 1-3). 
The adjacency matrix $W$ and node's weight $PW$ have been calculated in this scanning. 
%Then, we use the Equ~\ref{equ:pq} to calculate the transitive matrix $\pi$ in line 4.
Line 4 is to calculate the transitive matrix using the $2^{nd}$ transfer probability formula provided by Node2vec.
Line 6-7 introduces how to generate the word sequences $SC$ by traveling on the word co-occurrence graph. 
The node on the graph is scanned one by one, and the number of times in which each node is traversed as the rooted node is equal to $n\frac{PW_u}{\sum_{k}PW_k}$. 

%Function RandomWalkSampleProcess, in line10-13, describes the process of generating a word sequence by random walking. The alias sample is used to speed up the traveling process. 
%In the end, in line 8, the word embedding $\Phi$ is calculated by~\SG~based on the sampling word corpus $SC$. 

\subsection{Time and Space Complexity Analysis.}
\label{sec:time_complexity}

 Denoting N as the total corpus size and $V$ as the unique word vocabulary count.
% ,~\M~has a high time complexity in training word embedding with Skip-Gram, which is $O(Nlog(V))$. $N$ will be very big on large corpus (e.g., $N \approx 1.1e^{9}$ on 1GB data set, and $\approx 4.5e^{9}$ on 4GB data set). However, 
 In ~\M, the training corpus for Skip-Gram is our sampling corpus whose size is nl, where n is the number of random walks and l is the walk length. The time complexity is reduced to $O(nllog(V))$, since $n$l is smaller than $N$ ($nl\approx1.9e^{8}$ for both 1GB and 4GB data set).
% On large-scale data sets, $V$ is the fixed value, and $nl$ is also set by the user so that our method’s training time is relatively stable for various-size data sets.

As for the space complexity,~\M~needs to store the word graph, the generated word corpus, and the final word embedding.
So, the space complexity of the word graph is $O(m|V|)$, $m$ is the average degree of the graph; 
the space complexity of generated word corpus is $O(nl)$;
the word embedding is $O(d|V|)$.
Therefore,  the space complexity of~\M~is $O(m|V| + nl + d|V|)$. 
\section{Experimental Analysis}
\label{sec:experiment}

\subsection{\textbf{Experimental Setting}}
\label{sec:exp_set}

% In the experiment, Pecanpy model~\cite{liu2021pecanpy}, which is a new Python implementation of node2vec. Because of the poor scale and network density of the original \textit{Python} and \textit{C++} implementations of the node2vec model, it uses cache optimized compact graph data structure and pre-computing/parallelization to improve the shortcomings of node2vec. Thus, word embedding can be obtained faster together with higher quality.

% We first explored the impact of different parameter values on word embedding quality, compared two different word weight setting methods through experiments, and selected the best experimental parameter values and word weight setting methods for subsequent experiments.
%Then, we trained with word2vec and ~\M~on three corpora, evaluated and compared the word embedding quality on three tasks: word similarity, word analogy, and classification tasks.
%We recorded the time cost in each training, and generated two larger corpora for more in-depth time performance evaluation.

% \subsection{\textbf{Datasets}}
We use various data sets to test our approaches. The processed dataset information can be found in Table \ref{tab:data_statistic}.

\noindent \textbf{Text8}~\footnote{\url{http://mattmahoney.NET/dc/text8.zip}}: 
Text8~\cite{multimedia2009large} contains 100M processed Wikipedia characters created by changing the case to lower of the text and removing any character other than the 26 letters $a$ through $z$. Meanwhile, PyDictonary, as in the English language dictionary, Wordnet lexical database, and Enchant Spell Dictionary, is applied to filter the correct English words.
% After pre-processing, Text8 contains 135,317 various words, and 3,920,065 word co-occurrence relationships, and the density of the word co-occurrence graph is around 0.02\%. See Table \ref{tab:data_statistic} for the specific data of Text8.  

\noindent \textbf{One billion words benchmark (1b words banchmark)} ~\footnote{\url{https://www.kaggle.com/datasets/alexrenz/one-billion-words-benchmark}}: 
This is a new benchmark corpus with nearly 1 billion words of training data, which is used to measure the progress of statistical language modeling. 
%It was initially published in~\cite{chelba2013one}. The same pre-processing technique is applied to filter the data set. See Table \ref{tab:data_statistic} for the specific data of preprocessed 1b words benchmark.    

\noindent \textbf{Enlish Wikipedia data set (En-Wikipedia)}~\footnote{\url{https://dumps.wikimedia.org/enwiki/latest/enwiki-latest-pages-articles.xml.bz2}}: 
This is a word corpus of English articles collected from Wikipedia web pages. 
%See Table \ref{tab:data_statistic} for the specific data of preprocessed 1b words benchmark.   

\noindent \textbf{Concatenating data set:}  
To test the scalability of our approach,  we also process several En-Wikipedia data sets successively as a single sequential data set. We use Con-En-Wikipedia-i to denote the concatenating data set with $i\mbox{-}th$ data sets merged together. By the way, Con-En-Wikipedia-2 is 16.4G, and Con-En-Wikipedia-3 is 24.6 G.

\vspace{-0.3in} 
\begin{table}
	\tiny
	\caption{Statistics  of Datasets} 
	\label{tab:data_statistic}
	\resizebox{\linewidth}{!}{
		\begin{tabular}{|c|c|c|c|c|}
			\hline
			Data sets & Size & $ |\mathrm{V}| $ & $ |\mathrm{E}| $& Density \\ \hline
			\tiny Text8 & 95.3M & 135,317 & 3,920,065 & 0.02$\%$\\  \hline
			\tiny 1b words benchmark & 2.51G & 82,473 & 54,125,475  & 0.80$\%$\\  \hline
			\tiny En-Wikipedia & 8.22G & 100,723 & 64,633,532 & 0.64$\%$\\ \hline
		\end{tabular}
	}
\end{table}

\noindent \textbf{Baseline method:} 
% 需要在此处Baseline添加FastText相应内容
Word2vec and Fasttext are two standard methods for training static word embedding, so we use them as our baseline to compare with ~\M. For these two baselines, all parameters are the default values provided by the original function

All our experiments are conducted on a PC with a 1.60GHz Intel Core 5 Duo Processor, 8GB memory, and running Win10, and all algorithms are implemented in Python and C++.
Meanwhile, instead of using Node2vec, the Pecanpy model~\cite{liu2021pecanpy}, which uses cache optimized compact graph data structure and pre-computing/parallelization to improve the shortcomings of Node2vec, is applied in our experiment. 
%Because the word graphs we constructed are relatively sparse (the density is no more than 2$\%$, see Table~\ref{tab:data_statistic}), according to Pecanpy's distinction of operation modes of networks with different densities, $ SparseOTF $ mode is selected, which is suitable for networks that are large and sparse.

\subsection{\textbf{Evaluation Criteria}}
\label{sec:eval}
We adopted the three evaluation tasks mentioned in \cite{schnabel2015evaluation}: Categorization, Similarity, and Analogy, and we tested them with the method proposed in \cite{jastrzebski2017evaluate}.

\noindent \textbf{Categorization:} The goal here is to restore word clusters to different categories. Therefore, the corresponding word embedding of all words in the data set are clustered, and the purity of the cluster is calculated according to the marked data set. 
%The selected two evaluation data sets are Battig and BLESS. 
%The Battig comprises 83 concepts from 10 common concrete categories (up to 10 concepts per class), with the concepts selected so that they are rated as highly prototypical of the class. Class examples include land mammals (e.g., dog, elephant, etc.), tools (e.g., screwdriver, hammer, etc.), and fruit (e.g., orange, plum, etc.). The BLESS data set, designed for the evaluation of distributional semantic models, contains 200 distinct English concrete nouns as target concepts, which are categorized into 17 broad classes. 

\noindent  \textbf{Similarity:} 
%The two evaluation data sets (Simlex999 and MEN) are selected in our experiments. These data sets contain a list of association scores for paired words and a related human judgment similarity. The cosine similarity in two words should be highly correlated with the human correlation score.
This task requires calculating the cosine similarity of paired words calculated using word vectors and comparing it with the relevant human judgment similarity.

\noindent \textbf{Analogy:}The goal is to find a term $x$ for a given term $y$ so that $x:y$ is most like the sample relationship $a:b$. It requires predicting the degree to which the semantic relations between $x$ and $y$ are similar to those between $a$ and $b$.

The word similarity prediction effectiveness is measured with the help of Spearman's rank correlation coefficient $\rho$~\cite{myers2013research}.
%  It is an index to measure the dependence of two variables. If there are no duplicate values in the data and the two variables are completely monotonically correlated, the value of $\rho$ is +1 or $ \mbox{-} $1.This measures the rank correlation (higher is better) between the list of word pairs sorted in decreasing order of inter-similarity values predicted by a word embedding algorithm and the reference list of human-judged word pairs. 
For the analogy and the concept categorization tasks, we report the accuracy~\cite{metz1978basic} in predicting the reference word and that of the class, respectively~\footnote{The detailed information and the source code shows on this website~\url{ https://github.com/kudkudak/word-embeddings-benchmarks}}. 

%\subsection{\textbf{Intuitive Result}}
%
%In accuracy tasks, we do not pursue better than Word2vec or~\FT, but we can foresee that~\M~will not suffer a loss of accuracy because of the characteristic of the sampling method. 
%In time-consuming tasks, because the vocabulary and density of the graph are relatively stable, with the increase in data volumes, the runtime of~\M~increases slowly. 
%However, the runtime of Word2vec and~\FT~increase linearly with the increase in data volumes. 
%Therefore, in the beginning, the runtime of~\M~is much higher than Word2vec and~\FT~, but the runtime of Word2vec and~\FT~quickly rise noticeably with the increase in data volume.     

\subsection{\textbf{Parameters Study}}

\noindent \textbf{Study of P, Q:} 
We explored the best combination of parameters $ p $ and $ q $ on $ [(0.001, 1), (1, 1), (1, 0.001)] $. 
Table~\ref{tab:pq} shows the evaluation results of the three tasks. It can be seen from the results that the $ (p, q) $combination of $ (1, 0.001) $ can obtain better accuracy. 

\noindent \textbf{Node's weight study:} 
Table \ref{tab:pq} shows a comparison of the evaluation results of word embedding obtained by setting various word weight. 
Method 1 (Pecanpy+TF) uses $ TF $ value as the weight of the node;
Method 2 (Pecanpy+TF-IDF) uses  $ TF\mbox{-}IDF $ value;
Method 3 (Pecanpy) do not set the node's weight.
The results show that using TF-IDF to set weights is the best, so we will use this method in the following experiments. 

\vspace{-0.3in} 
\begin{table}[htb]
	\caption{Evaluation results of different $ p$ , $ q $ combinations and word weight setting}
	\newcommand{\tabincell}[2]{\begin{tabular}{@{}#1@{}}#2\end{tabular}} 
	\resizebox{\linewidth}{!}{
		\begin{tabular}{|c|c|c|c|c||c|c|c|}
			\hline		
			\multirow{2}{*}{Tasks}  & \multirow{2}{*}{Data set} & \multicolumn{3}{|c||}{Different $ (p, q) $ Combimations } & \multicolumn{3}{|c|}{Different  Word Weight Setting Method} \\ \cline{3-8}
			\multicolumn{1}{|l|}{} & & $ (0.001, 1) $ & $ (1, 1) $ & $ (1, 0.001) $ & Pecanpy+TF & Pecanpy+TF-IDF & Pecanpy\\ \hline
			\multirow{2}{*}{\tabincell{c}{Categorization Tasks \\ (Accuracy*100)}}  
			&BLESS             & 48.0    & 59.5    & \textbf{66.0}     & 60.5         & \textbf{61.0}    & 60.0 \\
			&Battig            & 25.1    &30.0     & \textbf{32.2}     & 32.1         & \textbf{34.3}    & 30.1 \\ \hline
			\multirow{2}{*}{\tabincell{c}{Word Similarity \\ (Spearman's $ \rho $ *100)}} 
			&MEN               & 37.7    & 52.4    & \textbf{56.0}     & \textbf{54.5}    & 54.4               & 54.1\\
			&SimLex999         & 20.6    &18.1     & \textbf{21.5}     & 21.3             & \textbf{21.7}      & 20.4 \\
			\hline
			\multirow{2}{*}{\tabincell{c}{Word Anology \\ (Accuracy(P@1)*100)}} 
			&MSR               & 4.0    & \textbf{15.7}    & 11.9      & \textbf{12.9}    & 12.3               &    12.1\\
			&SemEval2012\_2    & 8.4    & \textbf{13.5}    & 10.4      & 11.0             & \textbf{11.5}      &    10.3\\
			\hline
		\end{tabular}
	}
	\label{tab:pq}
\end{table}

%\begin{table}[htb]
%	\caption{Word Weight Setting Experiment Result}
%	\newcommand{\tabincell}[2]{\begin{tabular}{@{}#1@{}}#2\end{tabular}} 
%	\resizebox{\linewidth}{!}{
%		\begin{tabular}{|c|c|c|c|c|}
%			\hline		
%			\multirow{2}{*}{Tasks}  & \multirow{2}{*}{ Dataset} & \multicolumn{3}{|c|}{Method}\\ \cline{3-5}
%			\multicolumn{1}{|l|}{} & & Pecanpy+TF & Pecanpy+TF-IDF & Pecanpy\\ \hline
%			\multirow{2}{*}{\tabincell{c}{Categorization Tasks \\ (Accuracy*100)}}  
%			&BLESS         & 60.5         & \textbf{61.0}    & 60.0           \\
%			&Battig  & 32.1         & \textbf{34.3}    & 30.1              \\ \hline
%			\multirow{2}{*}{\tabincell{c}{Word Similarity \\ (Spearman's $ \rho $ *100)}} 
%			&MEN    & \textbf{54.5}    & 54.4               & 54.1        \\
%			&SimLex999  & 21.3             & \textbf{21.7}      & 20.4       \\
%			\hline
%			\multirow{2}{*}{\tabincell{c}{Word Anology \\ (Accuracy(P@1)*100)}} 
%			&MSR               & \textbf{12.9}    & 12.3               &    12.1     \\
%			&SemEval2012\_2    & 11.0             & \textbf{11.5}      &    10.3     \\
%			\hline
%		\end{tabular}
%	}
%	\label{fig:node_weight}
%\end{table}

Due to space limitations, we have only reported test results for some important parameters. Our final experiment used the best performing values for each parameter in the test.

\vspace{-0.1in} 
\subsection{\textbf{Accuracy Experiments}}

We compared the performance of~\M, Word2vec, and~\FT~on three tasks. 
Table~\ref{ex_result} show the experimental results of the categorization task, word similarity task and word analogy task. 
The experimental results show that: 
(i) On the three tasks, the quality of word embedding trained by~\M, Word2vec, and~\FT~will improve with the increased data set.
%%The larger the corpus is, the higher the quality of word embedding can be learned. 
%(ii) In general, the accuracy of~\M~on Text8 is slightly lower than Word2vec and~\FT.~\M's random walk process aggregates statistically frequently coexisting words into a sequence, which is equivalent to pre-learning the relationship between words. 
%If the corpus is too small, the walking process will focus on words with high probability while ignoring the possible impact of words with low probability. 
%This leads to the final sampling result not gathering all the words that may be semantically related in one sampling sequence. 
%(iii) The accuracy of~\M~on 1b words benchmark and En-Wikipedia is close to or higher than Word2vec and~\FT.
%With the increased corpus size, 
We can see that the performance of~\M~is gradually close to Word2vec and~\FT, or even better. 
%The sample size needed to produce accurate results in a study is often surprisingly small, followed by the central limited theorem (CLT) \cite{rosenblatt1956central}. 
This shows that a random walk has a certain probability to capture marginal words that appear relatively few times. Thus, this makes the sampled corpus closer to the real text and even can follow the similar word's distribution in advance. 
Moreover, this may be the reason to explain that the word embedding obtained by~\M~performs better in some test tasks.

\begin{table}[htb]
	\caption{Precision Experiments Results}
	\newcommand{\tabincell}[2]{\begin{tabular}{@{}#1@{}}#2\end{tabular}} 
	\resizebox{\linewidth}{!}{
		\begin{tabular}{|c|c|c|c|c|c|c|c|}
			\hline		
			\multicolumn{2}{|c|}{Tasks} & \multicolumn{2}{|c|}{Categorization Tasks} & \multicolumn{2}{|c|}{Word Similarity} & \multicolumn{2}{|c|}{Word Anology}\\ \hline
			
			\multirow{2}{*}{Datasets}  & \multirow{2}{*}{Method} & \multicolumn{2}{|c|}{Accuracy*100}& \multicolumn{2}{|c|}{Spearman's $ \rho $ *100} & \multicolumn{2}{|c|}{Accuracy(P@1)*100}\\
			
			\cline{3-8}\multicolumn{1}{|l|}{} & & BLESS & Battig & MEN & SimLex999 & MSR & SemEval2012\_2 \\ \hline	
			
			\multirow{2}{*}{Text8}  
			&  Word2vec   & 58.0 & 35.2 & \textbf{63.0}& 26.3& 27.6& \textbf{13.2}\\
			& \FT & 50.5 & \textbf{36.2} & 61.1 & \textbf{27.5} & \textbf{41.4} & 10.1\\
			& \M & \textbf{66.0} & 32.2 & 56.0 & 21.5 & 11.9 & 10.4\\  \hline
			
			\multirow{2}{*}{\tabincell{c}{ 1b Words \\ Benchmark}} 
			&  Word2vec  & 77.0 & 30.8 & 68.6 & \textbf{33.0} & 34.8& 13.9\\
			& \FT & 72.5 & \textbf{32.3} &70.0& 30.3& \textbf{37.5}&12.2\\ 
			& \M & \textbf{83.5} & 31.6 &\textbf{70.1}& 31.6& 24.8& \textbf{15.7}\\  \hline
			
			\multirow{2}{*}{En-Wikipedia} 
			&  Word2vec  & 70.5 & 35.4 &67.7& 28.6&30.8&13.8\\
			& \FT & 77.0 & \textbf{37.3} &\textbf{69.0}&28.4&\textbf{42.8}&15.6\\ 
			& \M & \textbf{83.5} & 36.5 &66.5&\textbf{29.7}&28.8&\textbf{15.9}\\ \hline
		\end{tabular}
	}
	\label{ex_result}
\end{table}

\subsection{\textbf{Efficiency of proposed algorithm}}

\begin{figure}[t]
	\centering
	\includegraphics[width=6cm,height=5cm]{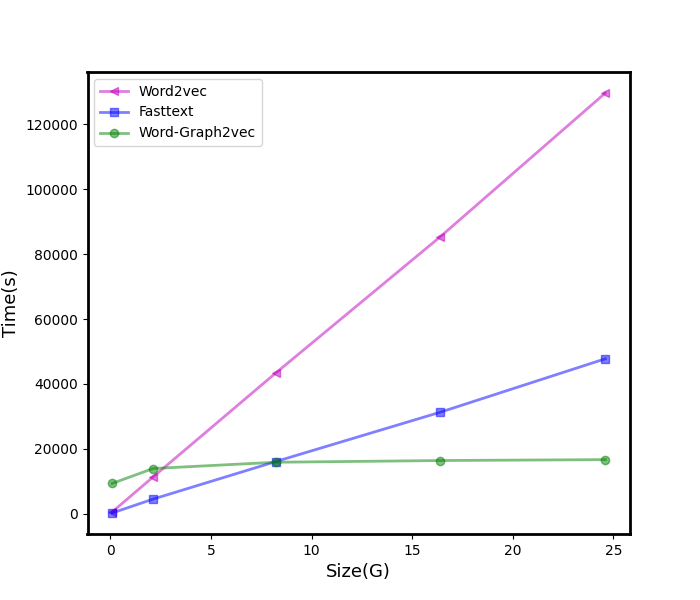}
	\caption{Times (h) vs. Size (G)}
	\label{fig:LINEC}
\end{figure}

In order to observe the time trend more intuitively, we generated two more extensive data sets (Con-En-Wikipedia-1, 16.4G; Con-En-Wikipedia-2, 24.6G) for testing using the En-Wikipedia. 
As shown in Figure~\ref{fig:LINEC}, we have plotted the experimental results on four data sets of different sizes (2.51G, 8.22G, 16.4G, 24.6G) into time curves (including 1b Words Benchmark and En-Wikipedia). 
Figure~\ref{fig:LINEC} indicates that the runtime of Word2vec and \FT~increase almost linearly. But~\M~'s runtime increases slowly (the slight increase of runtime is due to loading the word corpus). 
%The reason can be explained as no matter how large the original corpus is, the number of word sequence samples is a constant, so the training time of~\M~grows slightly.

\section{Conclusion}
\label{sec:conclusion}

We propose the~\M~algorithm to improve the performance of word embedding, which converts the large corpus into a word co-occurrence graph, then takes the word sequence samples from this graph by randomly walking and trains the word embedding on these sampling corpora in the end. 
We argue that due to the stable vocabulary, relative idioms, and fixed expressions in English, the size and density of the word co-occurrence graph change slightly with the increase of the training corpus. 
Thus,~\M~has a stable runtime on the large-scale data set, and its performance advantage becomes more and more obvious with the growth of the training corpus.
Experimental results show that the proposed algorithm outperforms traditional Word2vec in terms of efficiency and two-three times than~\FT.
\section{Acknowledgment}
\label{sec:Acknowledgment}

The authors acknowledge funding from 2023 Special Fund for Science and Technology Innovation Strategy of Guangdong Province (Science and Technology Innovation Cultivation of College Students), the Shenzhen High-Level Hospital Construction Fund (4001020), and the Shenzhen Science and Technology Innovation Committee Funds (JSGG20220919091404008, JCYJ20190812171807146).

\bibliography{references}
\bibliographystyle{splncs04}

\end{document}